\documentclass[final,5p,times,twocolumn]{elsarticle}
\usepackage{xcolor}

\usepackage{amssymb}
\usepackage{amsmath}
\usepackage{amsfonts}
\usepackage{physics}
\usepackage[inline]{showlabels}
\date{January 20, 2024}

\journal{Chaos, Solitons \& Fractals}

\begin{document}

\begin{frontmatter}

\title{Impact of white Gaussian internal noise on analog echo-state neural networks}

\author[inst1]{Nadezhda Semenova}
\ead{semenovni@sgu.ru}

\affiliation[inst1]{organization={Institute of Physics, Saratov State University},
            addressline={Astrakhanskaya str., 83}, 
            city={Saratov},
            postcode={410012}, 
            country={Russia}}           

\begin{abstract}
In recent years, more and more works have appeared devoted to the analog (hardware) implementation of artificial neural networks, in which neurons and the connection between them are based not on computer calculations, but on physical principles. Such networks offer improved energy efficiency and, in some cases, scalability, but may be susceptible to internal noise. This paper studies the influence of noise on the functioning of recurrent networks using the example of trained echo state networks (ESNs). The most common reservoir connection matrices were chosen as various topologies of ESNs: random uniform and band matrices with different connectivity. White Gaussian noise was chosen as the influence, and according to the way of its introducing it was additive or multiplicative, as well as correlated or uncorrelated. In the paper, we show that the propagation of noise in reservoir is mainly controlled by the statistical properties of the output connection matrix, namely the mean and the mean square. Depending on these values, more correlated or uncorrelated noise accumulates in the network. We also show that there are conditions under which even noise with an intensity of $10^{-20}$ is already enough to completely lose the useful signal. In the article we show which types of noise are most critical for networks with different activation functions (hyperbolic tangent, sigmoid and linear) and if the network is self-closed.
\end{abstract}




\end{frontmatter}


\section*{Introduction}
Artificial neural networks (ANNs) are the basis of modern artificial intelligence and machine learning. They are able to learn, recognize hidden dependencies, and classify or predict with amazing accuracy. Their structure and overall operation were originally inspired by biological neural networks, and the first simple learning networks appeared several decades ago. But now, with the increasing performance of modern computers, there is a huge growth in various interesting applications of neural networks and topologies. Nowadays, they are successfully used in medicine \cite{Amato2013, Salahuddin2022, Sarvamangala2022, Celard2023}, finance \cite{Lazcano2023, Li2023}, technology \cite{Aghbashlo2015, Almonacid2017, Yang2023}, biology \cite{Marabini1994, Suzuki2011, Samborska2014}, and for predicting the behavior of complex, chaotic systems \cite{Wang2024}.

Recurrent neural networks (RNNs) play a key role in the field of machine learning when it comes to problems involving sequences of data, such as natural language or time series. They are able to capture temporal dependencies in data due to their unique architecture, which allows information to circulate within the network over several time steps \cite{Salehinejad2018}. This makes RNNs exceptionally suitable for tasks such as speech recognition \cite{Graves2014}, joint language and translation modeling \cite{Auli2013} and sequence prediction.

In this paper we study how noise affects one of the simplest RNNs, echo state network (ESN) \cite{Jaeger2001}. ESNs consists of one hidden layer called reservoir, input and output neurons. The states of neurons inside reservoir are determined by the input signal and the previous states of reservoir neurons.

Complex tasks such as large data processing or real-time data processing require significant amounts of computing resources and energy. Therefore, problems with energy efficiency, speed and scalability of ANNs are some of the main limitations of their application in various fields \cite{Markovic2020}. In addition, augmenting the network to find complex patterns can lead to problems with scalability and computing resource management \cite{Christensen2022}.

To overcome these limitations, there is active research into developing more energy-efficient neural network architectures, optimizing computation and developing specialized hardware to perform neural network operations, called hardware neural networks \cite{Seiffert2004, Misra2010, Bouvier2019}. Such efforts are aimed at improving the efficiency and performance of artificial neural networks and open the prospect of their wider application in various fields. In hardware neural networks, the artificial neurons and connection between them are based on physical principles such as optical neural networks \cite{Wang2022, Ma2023} memristive devices \cite{Tuma2016,Lin2018,Xia2019}, spin-torque oscillators \cite{Torrejon2017}, Mach–Zehnder interferometer \cite{Shen2017, Cem2023}, photoelectronic \cite{Chen2023}, coherent silicon photonics \cite{MourgiasAlexandris2022} and etc.

Noise in hardware neural networks can arise from a variety of sources, such as analog noise in digital device circuits, supply voltage changes, state transitions, and other analog and digital noise. This noise can have a negative impact on the accuracy of the network, as it can cause errors in the transmission and processing of information. There are many works dealing with noise in the input signal of ANNs \cite{Maas2012, Burger2012, Seltzer2013}, but in the case of hardware ANNs we are talking about internal noise inside the network. There are several papers that describe the properties of various internal noise sources in hardware ANNs \cite{Dolenko1993, Dibazar2006, Soriano2015, Janke2020, Nurlybayeva2022, Ma2023}. 

In our previous studies we have explained how white Gaussian noise affects deep feedforward neural networks \cite{Semenova2022NNs} in simplified and trained cases. Then we considered the impact of noise on networks which are similar to ESNs but very simplified and untrained \cite{Semenova2023ND, Semenova2023PND}. This paper complicates the task by considering real trained networks with different types of connection within a reservoir. The noise intensities and properties came from photonic recurrent neural network realized in hardware \cite{Bueno2018}. Therefore, here we consider additive and multiplicative, correlated and uncorrelated noise.

This paper starts explaining the considered ESN (Sec.~\ref{sec:system_esn}) and noise types (Sec.~\ref{sec:system_noise}). Then we explain how additive and multiplicative noise affects the output layer of ESN and how variance (dispersion) and error of the output signal is changed for different noise intensities (Sec.~\ref{sec:noise_output_layer}). Section \ref{sec:noise_reservoir} discusses how all four noise types changes the output signal if the noise is introduced into reservoir. For both cases, we provide not only the results of numerical simulation, but also the results of analytical prediction based on methods described in Sec.~\ref{sec:analytics}. Finally, we show what to expect if ESN in self-closed (Sec.~\ref{sec:closed_ESN}). It is also worth noting that we mostly use the hyperbolic tangent in the paper as the activation function for the reservoir neurons, but in Sec.~\ref{sec:activation_funciton} we show what changes if we consider a linear function or sigmoid.

\section{System under study}\label{sec:system}
\subsection{Echo-state neural network}\label{sec:system_esn}
\begin{figure}[h]
\includegraphics[width=\linewidth]{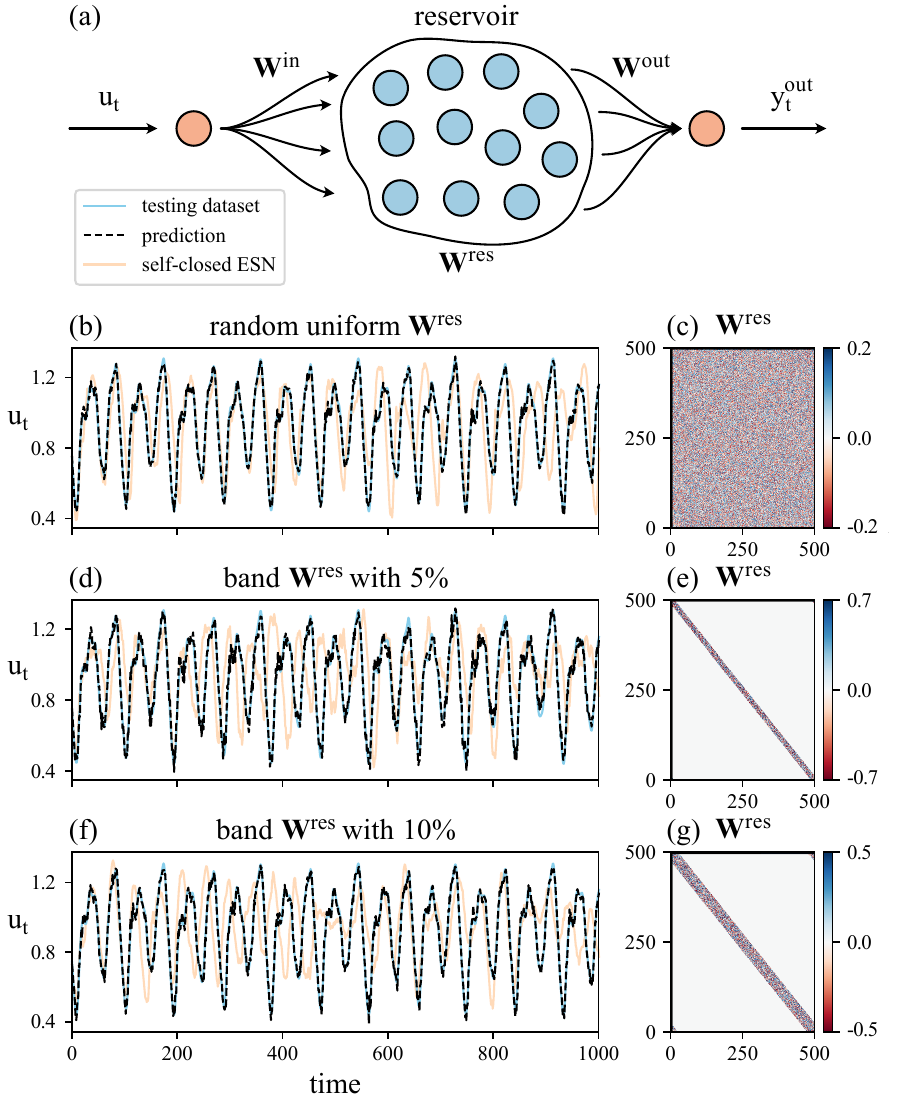}
\caption{Schematic representation of considered ESN (a) and how it can be used to predict chaotic time series. Panels (b,d,f) show the original time series (blue), prediction of ESN (black) and prediction of self-closed ESN (orange). These panels were prepared for three trained ESNs with random uniform $\mathbf{W}^\mathrm{res}$ (b,c) and band matrices with $5\%$ connectivity (d,e) and $10\%$ connectivity (f,g).}
\label{fig:scheme}
\end{figure}

In this paper we consider the impact of different sources of white Gaussian noise on echo-state neural network (ESN) schematically shown in Fig.~\ref{fig:scheme}(a). This network is the simplest example of a recurrent neural network, where the time-delayed feedback is realized via a reservoir with artificial neurons. From the mathematical standpoint, the signal of each neuron inside reservoir is determined by the input signal of the network $u_t$ at time $t$ and the signals coming from reservoir neurons $\mathbf{y}^\mathrm{res}_{t-1}$ at the previous time $(t-1)$:
\begin{equation}\label{eq:x_res}
\mathbf{x}^\mathrm{res}_t = f\big( u_t\mathbf{W}^\mathrm{in} + \mathbf{y}^\mathrm{res}_{t-1}\mathbf{W}^\mathrm{res}\big),
\end{equation}
where $\mathbf{x}^\mathrm{res}_t$ is the vector $(1\times N)$ determining the state of each neuron in reservoir before noise impact, $N=500$ is the number of neurons in reservoir. Connection between neurons is determined through the matrices connecting input neuron with reservoir $\mathbf{W}^\mathrm{in}$ of size $(1\times N)$ and reservoir-to-reservoir matrix $\mathbf{W}^\mathrm{res}$ of size $(N \times N)$. The function $f(\cdot)$ is the activation function. Here we will mainly use the hyperbolic tangent function $f(x)=\tanh(x)$. This function may be changed only in Sec.~\ref{sec:activation_funciton}, where we are going to consider the impact of activation function on obtained results.

Further, we will study the impact of noise on reservoir, and then the noisy output of reservoir neurons can be described using noise operator $\hat{\mathbf{N}}$ which meaning will be described in detail in Sec.~\ref{sec:system_noise}:
\begin{equation}\label{eq:Noise_operator}
\mathbf{y}^\mathrm{res}_t = \hat{\mathbf{N}} \mathbf{x}^\mathrm{res}_t.
\end{equation}
Further, as it has been already introduced for (\ref{eq:x_res}), all variables before or without noise impact are marked by $x$ letter, while $y$ is used for the values after applying noise. If at some stage the noise is turned off, we will assume that $y^\mathrm{res}_{t,i}=x^\mathrm{res}_{t,i}$, where $i$ is the number of neuron in reservoir, or $y^\mathrm{out}_{t}=x^\mathrm{out}_{t}$ for the output layer.

The output neuron of ESN is connected to the reservoir via the connection matrix $\mathbf{W}^\mathrm{out}$ of size $(N\times 1)$, and the output of ESN at time $t$ is therefore
\begin{equation}\label{eq:x_out}
x^\mathrm{out}_t = \mathbf{y}^\mathrm{res}_t \mathbf{W}^\mathrm{out}; \ \ \ \ y^\mathrm{out}_t = \hat{\mathbf{N}} x^\mathrm{out}_t.
\end{equation}

It is important to note, that in this paper we are mainly focused on the influence of noise on trained ESNs. Therefore, we need to describe in more detail how the networks were trained. The peculiarity of training an echo network is that during the training process only one connection matrix $\mathbf{W}^\mathrm{out}$ changes, and the rest are set at the beginning and then do not change at all. There are two options for how the connection matrix $\mathbf{W}^\mathrm{in}$ can be specified. One option is random real numbers in the range $[-1; 1]$ or $[-0.5; 0.5]$. This increases the complexity of the network, but makes the learning process less predictable. On the other hand, the matrix filled with units leads to more predictable results, and the propagation of noise in a self-closed ESN will be more understandable. Therefore, here we use the last case: $W^\mathrm{in}_i=1$, where $i=1,\dots N$. 

As for the reservoir matrix $\mathbf{W}^\mathrm{res}$, there are also several ways to specify it, and in this work we will focus on the two most commonly used types: 1 -- random uniform and 2 -- band matrix with determined spectral radius $s$. The uniform connection matrix is set randomly with uniform distribution from -0.5 to 0.5. To get the matrix with specified spectral radius $s$ we need to divide it by its spectral radius and multiply by $s$, so the boundaries of original distribution are not important. The example of such connection matrix is given in Fig.~\ref{fig:scheme}(c). The band matrix is a matrix whose non-zero elements are confined to a diagonal band, comprising the main diagonal and zero or more diagonals on either side. The number of non-zero elements is controlled by the bandwidth parameter, but here we will use $k$\% meaning the percentage of connected neurons. The non-zero values are set randomly in the same way as in uniform random matrix, then the matrix $\mathbf{W}^\mathrm{res}$ is multiplied by some value to get a specified spectral radius $s$ as was for uniform matrix. Figure \ref{fig:scheme}(e,g) show two band matrices with 5\% and 10\% connectivity, respectively. 

To train the network and to get the values of matrix $\mathbf{W}^\mathrm{out}$ connecting the reservoir at time $t$ with the output neuron we use pseudo-inverse matrix and Ridge regression. comparing ESN's output $y^\mathrm{out}_t$ with known value $u_{t+1}$. 

In order to train the network, we use the common task of predicting the time series of Mackey–Glass equation:
\begin{equation}\label{eq:MG}
\frac{du}{dt} = \beta\frac{u_\tau}{1+u^n_\tau}-\gamma u,
\end{equation}
where $u=u(t)$ is the system state at time $t$, $u_\tau=u(t-\tau)$ is its state at delayed time $(t-\tau)$. The rest parameters are $\beta=0.2$, $\gamma=0.1$, $\tau=16$, $n=10$. The system (\ref{eq:MG}) was integrated by Euler method with unity time step. In order to train the ESN we use the time series of system (\ref{eq:MG}) for $t\in[0;40000]$ as the input signal $u_t$ of ESN. Since the prediction problem is being solved, the same implementation is used as the correct output signal $y^\mathrm{out}_t$, but for the interval $t\in [1;40001]$.

For testing we use the same system, but at time $t\in[40000;41000]$. The true original realization is show by the blue lines in Figs.~\ref{fig:scheme}(b,d,f), the predicted outputs of ESN are given in black, while the output of self-closed ESN where the ESN's output is used as its input at the next time is plotted in orange. These results were obtained for connection matrices $\mathbf{W}^\mathrm{res}$ displayed on the right in the same Fig.~\ref{fig:scheme}. The statistical characteristics of these three matrices $\mathbf{W}^\mathrm{res}$ are given in Tab.~\ref{tab:W_res}, while the statistics of corresponding matrices $\mathbf{W}^\mathrm{out}$ after training can be found in Tab.~\ref{tab:W_out}

\begin{table}\begin{center}
\begin{tabular}{| c | c | c | c |}
\hline
$\mathbf{W}^\mathrm{res}$ & uniform & band 5\% & band 10\%  \\ \hline
$\mu(\mathbf{W}^\mathrm{res})$ & $-5.33\cdot 10^{-5}$ & $-0.0002$ & $0.0003$ \\
$\mu^2(\mathbf{W}^\mathrm{res})$ & $2.84\cdot 10^{-9}$ & $4.74\cdot 10^{-8}$ & $1.15\cdot 10^{-7}$ \\
$\eta(\mathbf{W}^\mathrm{res})$ & $0.0093$ & $0.0099$ & $0.0085$ \\
std$(\mathbf{W}^\mathrm{res})$ & $0.0965$ & $0.0998$ & $0.0923$ \\
spectral radius, $s$ & $2.2$ & $2.5$ & $2.2$ \\
\hline
\end{tabular}
\end{center}
\caption{Statistical characteristics of matrices $\mathbf{W}^\mathrm{res}$: mean value $\mu$, square of the mean value $\mu^2$, mean square value $\eta$, standard deviation and spectral radius.}
\label{tab:W_res}
\end{table}

\begin{table}\begin{center}
\begin{tabular}{|c | c | c | c |}
\hline
$\mathbf{W}^\mathrm{res}$ & uniform & band 5\%  & band 10\% \\ \hline
$\mu(\mathbf{W}^\mathrm{out})$ & $0.0041$ & $0.0048$ & $0.0044$ \\
$\mu^2(\mathbf{W}^\mathrm{out})$ & $1.69\cdot 10^{-5}$ & $2.31\cdot 10^{-5}$ & $1.92\cdot 10^{-5}$ \\
$\eta(\mathbf{W}^\mathrm{out})$ & $0.0004$ & $6.02\cdot 10^{-5}$ & $0.0007$ \\
std$(\mathbf{W}^\mathrm{out})$ & $0.0195$ & $0.0061$ & $0.0261$ \\
\hline
\hline
testing MSE & $0.0004$ & $0.0013$ & $0.0006$ \\
self-closed MSE & $0.0542$ & $0.0674$ & $0.0603$ \\
\hline
\end{tabular}
\end{center}
\caption{Statistical characteristics of matrices $\mathbf{W}^\mathrm{out}$ after training: mean value $\mu$, square of the mean value $\mu^2$, mean square value $\eta$ and standard deviation. The ESN performance are displayed by MSE calculated for predicted time series and time series of self-closed ESN.}
\label{tab:W_out}
\end{table}

\subsection{Noise in ESN}\label{sec:system_noise}
In this paper we consider the impact of white Gaussian noise on reservoir and output layer of ESN. The similar problem but for deep neural network was studied in our works \cite{Semenova2022NNs, Semenova2024}. These noise types and noise intensities were taken from hardware optical setup of ESN \cite{Bueno2018}. In this paper we consider the same types of noise introducing. We consider two types of noise depending on how the noise affects the neuron's output:
\begin{itemize}
\item additive noise: $y_{t,i} = x_{t,i} + \sqrt{2D^U_A}\xi^{UA}_{t,i}$ \\
\item multiplicative noise: $y_{t,i} = x_{t,i} \cdot\Big(1+ \sqrt{2D^U_M}\xi^{UM}_{t,i} \Big)$.
\end{itemize}
Here $\xi_{t,i}$ is white Gaussian noise source with zero mean. The parameter $D$ is the noise intensity, then the variance (dispersion) of this noise source is $\mathrm{Var}\Big[\sqrt{2D^U_A}\xi^{UA}_{t,i}\Big] = 2D^U_A$. The designations `A' and `M' correspond to additive and multiplicative noise, respectively. 

The indices $t$ and $i$ mean that the noise values vary over time $t$ and are different for each neuron $i$ from the same layer. The other two types of noise differs in the way of influence on a group of neurons:
\begin{itemize}
\item uncorrelated noise: $y_{t,i} = x_{t,i} + \sqrt{2D^U_A}\xi^{UA}_{t,i}$ \\
\item correlated  noise: $y_{t,i} = x_{t,i} + \sqrt{2D^C_A}\xi^{CA}_{t}$.
\end{itemize}
Here letters `C' indicate correlated noise producing the same noise value inside one layer (reservoir in this paper), and therefore it does not depend on neuron index $i$. 

Finally, we get four combinations of all noise types: 1 -- uncorrelated additive $\xi^{UA}_{t,i}$, 2 -- uncorrelated multiplicative $\xi^{UM}_{t,i}$, 3 -- correlated additive $\xi^{CA}_{t}$, 4 -- correlated multiplicative $\xi^{CM}_{t}$. The noise operator including all four noise types inside reservoir is
\begin{equation}\label{eq:noise_reservoir}
\begin{array}{c}
y^\mathrm{res}_{t,i} = \hat{\mathbf{N}} x^\mathrm{res}_{t,i} = x^\mathrm{res}_{t,i}\cdot\Big( 1+\sqrt{2D^U_M}\xi^{UM}_{t,i} \Big)\Big( 1+\sqrt{2D^C_M}\xi^{CM}_{t} \Big) + \\
\sqrt{2D^U_A}\xi^{UA}_{t,i} + \sqrt{2D^C_A}\xi^{CA}_{t}.
\end{array}
\end{equation}
The output layer of ESN consists of only one neuron and therefore, it makes no sense to separate the noise into correlated and uncorrelated noise for this layer. To maintain generality, we will continue to refer to them as correlated noise in equations:
\begin{equation}\label{eq:noise_out}
y^\mathrm{out}_{t} = \hat{\mathbf{N}} x^\mathrm{out}_{t} = x^\mathrm{out}_{t}\cdot\Big( 1+\sqrt{2D^C_M}\xi^{CM}_{t} \Big) + \sqrt{2D^C_A}\xi^{CA}_{t}.
\end{equation}

\section{Noise in the output layer}\label{sec:noise_output_layer}
\begin{figure}[h]
\includegraphics[width=\linewidth]{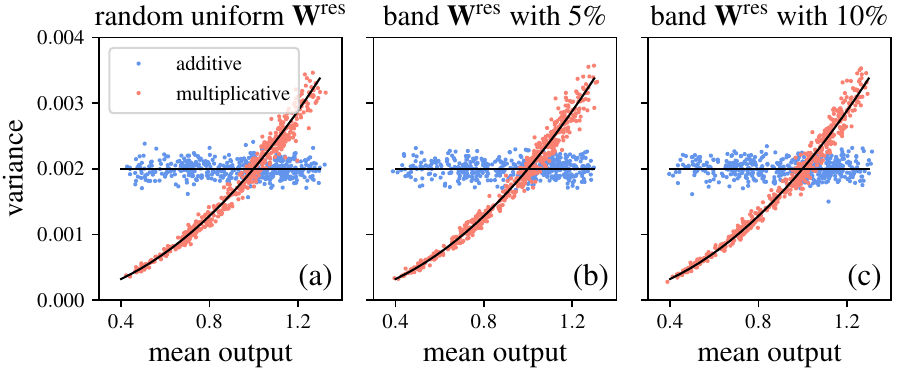}
\caption{Variance of the output ESN signal $y^\mathrm{out}_t$ in the case of noise in the output layer for three trained ESNs with random uniform $\mathbf{W}^\mathrm{res}$ (a), band matrix with 5\% connectivity (b) and band matrix with 10\% connectivity (c). Blue points correspond to additive noise with $D^C_A=0.001$, while red points were prepared for multiplicative noise with $D^C_M=0.001$. The analytics estimation of the variance level (black lines) were obtained using (\ref{eq:var_out}).}
\label{fig:noise_out_variance}
\end{figure}

In this section we consider the impact of additive and multiplicative noise on the output layer of trained ESN. As the first step, the ESN is not self-closed, so the input signal $u_t$ is noise free. Figure \ref{fig:noise_out_variance} illustrates the variance (dispersion) of ESN output  depending on its mean value prepared for all three trained networks.

Here we consider additive (blue points) and multiplicative (red points) noise separately. In our previous works we found that when considering both noise types, we get their superposition in variance. The points in Fig.~\ref{fig:noise_out_variance} are the result of numerical simulation. Here we used the same input time series as in Fig.~\ref{fig:scheme} containing $T=1000$ time points $u_t$. Each input signal $u_t$ is repeated $K=300$ to get the statistics of variance and expected (mean) value of the output signal $y^\mathrm{out}_t$. Therefore, Fig.~\ref{fig:noise_out_variance} contains $1000$ points for each noise type.

As it can be seen from Fig.~\ref{fig:noise_out_variance}, the noise inserted into the output layer does not depend on connection matrices, and the law describing the statistic of ESN output signal is the same. The range of mean output values is the same as the range of considered input values $u_t$. If one need to shift the points along the curves, then bias can be used in the output neuron.

The black lines in Fig.~\ref{fig:noise_out_variance} are the analytical predictions of the noise level obtained by the same technique that has been used in our paper \cite{Semenova2022NNs} for deep neural network. Its application for the output layer of ESN is given in Sec.~\ref{sec:analytics_output} in detail.

The variance of the output signal $\mathrm{Var}[y^\mathrm{out}_t]$ depending on its expected value (mean output) $\mathrm{E}[y^\mathrm{out}_t]$ in the case of noise in the output layer of ESN can be obtained as follows.
\begin{equation}\label{eq:var_out}
\mathrm{Var}[y^\mathrm{out}_t] = 2D^C_A + 2D^C_M\mathrm{E}^2[y^\mathrm{out}_t] + (1+2D^C_M)\mathrm{Var}[x^\mathrm{out}_t],
\end{equation}
where $\mathrm{E}[\cdot]$ is the expected value, and $\mathrm{E}^2[\cdot]$ means the square of this value. In the case of not self-closed ESN, when $u_t$ is taken from testing time series, not from ESN output $\mathrm{Var}[x^\mathrm{out}_t]=0$. Therefore, the variance level for additive noise in the output layer does not depend on the output signal being constant equal to $2D^C_A$. In the case of multiplicative noise in the output layer, the dependency of variance of the output ESN signal on its expected value is quadratic with a factor of $2D^C_M$. Analytical predictions of the noise level in Fig.~\ref{fig:noise_out_variance} are based on the law $g(x)= 2D^C_A+2D^C_M\cdot x^2$.

\begin{figure}[h]
\includegraphics[width=\linewidth]{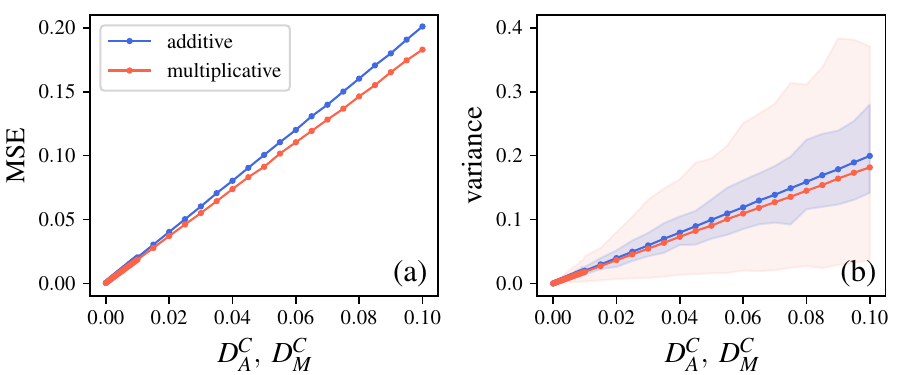}
\caption{Averaged MSE (a) and variance (b) of the output ESN signal $y^\mathrm{out}_t$ in the case of noise in the output layer of trained ESNs with uniform $\mathbf{W}^\mathrm{res}$. Blue color correspond to additive noise, while all dependences for multiplicative noise are shown in red. Coloured backgrounds in (b) show the ranges of variance.}
\label{fig:noise_out_mse}
\end{figure}

Figure~\ref{fig:noise_out_variance} was prepared for only one value of noise intensity $10^{-3}$. This value came from our observations of experimental photonic ESN \cite{Bueno2018}. But since in this paper we are interested in generality, Fig.~\ref{fig:noise_out_mse} shows how mean squared error (MSE) and variance change depending on the noise intensity. Figure~\ref{fig:noise_out_mse} was prepared for only one trained network (uniform random matrix), but for the other two the result was the same. Both dependencies MSE and variance are linear, an increase in noise intensity leads to linear enlargement of the output signal's variance and error. This is in a good correspondence with analytics (\ref{eq:var_out}).

This case is similar to the impact of noise on only one neuron with linear activation function. This case has been described in our paper \cite{Semenova2023ND}. At the same time, it is also important to note that in the case of self-closed ESN, when the output noisy signal becomes its input signal at the next time, the output noise can accumulate a lot. This case will be considered in Sec.~\ref{sec:closed_ESN}. 

\section{Noise in reservoir}\label{sec:noise_reservoir}
\begin{figure}[h]
\includegraphics[width=\linewidth]{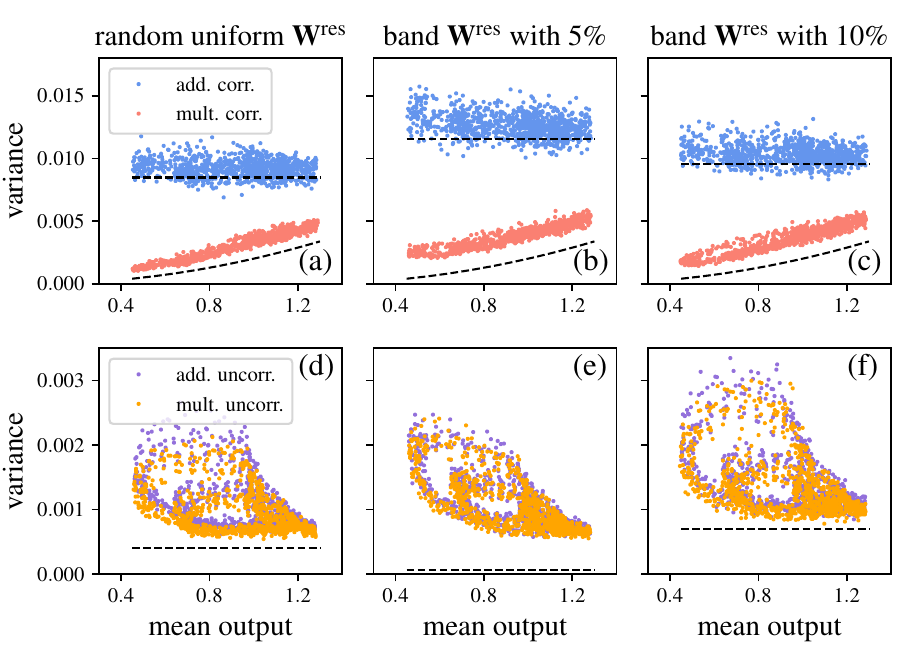}
\caption{Variance of the output ESN signal $y^\mathrm{out}_t$ in the case of noise in reservoir for three trained ESNs with different $\mathbf{W}^\mathrm{res}$. Four noise sources were considered: correlated additive (blue) and multiplicative (red); uncorrelated additive (purple) and multiplicative (orange). Black dashed lines show the analytical prediction of variance levels based on (\ref{eq:var_corr_noise_res}). All noise intensities are the same $D^U_A=D^U_M=D^C_A=D^C_M=10^{-3}$.}
\label{fig:noise_res_variance}
\end{figure}

In this section we study the influence of uncorrelated and correlated noise inside reservoir. Figure \ref{fig:noise_res_variance} shows the impact of correlated noise (top panels) and uncorrelated noise (bottom panels) inside reservoir on the output signal of ESN.

In the case of a noisy reservoir, more complex noise accumulation occurs even if the network is not self-closed. Analysis of the distribution of points and the shape of the dependence of variance on mean output signal does not provide a complete picture. Therefore, further we will consider both the results of numerical modeling and the results of analytical assessment to understand which network components produce more noise and which produce less. More detailed description of how the final equations were obtained is given in Sec.~\ref{sec:analytics_reservoir}.

The equation below includes all four considered noise sources inside reservoir.
\begin{equation}\label{eq:var_noise_res}
\begin{array}{c}
\mathrm{Var}[y^\mathrm{out}_t] = \mathrm{Var}[x^\mathrm{out}_t] = \mathrm{Var}[\sum\limits_j W^\mathrm{out}_j y^\mathrm{res}_{t,j}] = \\
\mathrm{Var}\Big[ \sqrt{2D^C_A}\xi^{AC}_t\cdot\sum\limits_j W^\mathrm{out}_j + \sum\limits_j W^\mathrm{out}_j\sqrt{2D^U_A}\xi^{AU}_{t,j} + \\
\big(1+\sqrt{2D^C_M}\xi^{MC}_t\big)\cdot\sum\limits_j W^\mathrm{out}_j x^\mathrm{res}_{t,j} \big(1+\sqrt{2D^U_M}\xi^{MU}_{t,j}\big) \Big] \approx \\
2D^C_A N^2\mu^2(\mathbf{W}^\mathrm{out}) + 2D^U_A N\eta(\mathbf{W}^\mathrm{out}) + 2D^C_M \mathrm{E}^2[y^\mathrm{out}_t] + \\
2D^U_M(1+2D^C_M)\cdot\sum\limits_j\Big( W^\mathrm{out}_j \mathrm{E}[x^\mathrm{res}_{t,j}] \Big)^2 + \\
(1+2D^U_M)(1+2D^C_M)N^2\mu^2(\mathbf{W}^\mathrm{out})\cdot\overline{\mathrm{Var}[\mathbf{x}^\mathrm{res}_t]},
\end{array}
\end{equation}
where $\mu^2(\mathbf{W}^\mathrm{out})$ is the square of mean value calculated for connection matrix $\mathbf{W}^\mathrm{out}$, $\eta(\mathbf{W}^\mathrm{out})$ is mean square value of this matrix. The term $\overline{\mathrm{Var}[\mathbf{x}^\mathrm{res}_t]}=\frac{1}{N}\sum_j \mathrm{Var}[x^\mathrm{res}_{t,j}]$ is the mean variance of variables $x$ averaged over $N$ neurons in the reservoir. These variables are the output signals of reservoir neuron before the noise impact at time $t$, but it indirectly includes noises that affected the reservoir at previous time steps. For now we will not focus on this term, but later it will be seen that it has a significant influence on the final noise level.

\subsection{Correlated noise}
The impact of correlated noise inside reservoir observed in numerical simulation is shown in Fig.~\ref{fig:noise_res_variance}(a--c) by blue points for correlated additive noise and by red points for correlated multiplicative noise. These three panels were prepared for three trained ESNs with different reservoir connection matrices. 

For correlated noise, it can be seen that its effect on the reservoir is very similar to its effect on the output layer. But from a quantitative point of view, the difference is quite noticeable, especially for additive correlated noise. In order to understand where this effect comes from, let us briefly consider the analytical assessment of the variance in the ESN output signal. By analyzing Eq.~(\ref{eq:var_noise_res}), we can write down only the terms that include correlated noise:
\begin{equation}\label{eq:var_corr_noise_res}
\begin{array}{c}
\mathrm{Var}[y^\mathrm{out}_t] = 2D^C_A N^2\mu^2(\mathbf{W}^\mathrm{out}) +  2D^C_M \mathrm{E}^2[y^\mathrm{out}_t] + \\
(1+2D^C_M)N^2\mu^2(\mathbf{W}^\mathrm{out})\cdot\overline{\mathrm{Var}[\mathbf{x}^\mathrm{res}_t]}.
\end{array}
\end{equation}
The last term in (\ref{eq:var_corr_noise_res}) shows the impact of noise inside reservoir from previous time moments. Black dashed lines show the analytical prediction of noise level based on (\ref{eq:var_corr_noise_res}) without the last term. The first term in (\ref{eq:var_corr_noise_res}) includes only additive correlated noise. As was for additive noise in the output layer, this term does not depend on expected value of the output signal, and therefore the noise level is constant. For noise in the output layer this constant was determined only by the noise intensity, but for the case of reservoir additive correlated noise, it also includes the statistical characteristics of the output connection matrix, namely $N^2\mu^2(\mathbf{W}^\mathrm{out})$. The horizontal black dashed lines in Fig.~\ref{fig:noise_res_variance} show the first term of (\ref{eq:var_corr_noise_res}) based on statistics of $\mathbf{W}^\mathrm{out}$  for all three trained matrices (see Tab.~\ref{tab:W_out}). 

Multiplicative correlated noise in reservoir at time $t$ can be roughly described by the second term in (\ref{eq:var_corr_noise_res}). This term is similar to what was observed for noise in the output layer. This term includes only the noise intensity $D^C_M$ and the square of expected output value $\mathrm{E}^2[y^\mathrm{out}_t]$ without any statistical characteristics of connection matrices. The noise level prediction based on the second term of Eq.~\ref{eq:var_corr_noise_res} is shown by the sloping black dashed lines in Fig.~\ref{fig:noise_res_variance}. As it was for additive correlated noise, the unaccounted last term in (\ref{eq:var_corr_noise_res}) clearly gives an almost constant addition to the final variance.

The analytical prediction for additive and multiplicative correlated noise is in good correspondence with numerical simulation, but there is definitely some additional constant shift coming from the last term with $\overline{\mathrm{Var}[\mathbf{x}^\mathrm{res}_t]}$. This term has a factor $N^2\mu^2(\mathbf{W}^\mathrm{out})$, and according to Tab.~\ref{tab:W_out}, this factor is the largest for the case of 5\%-band matrix and the smallest for uniform matrix. Moreover, the discrepancy between the numerical simulation and the dashed lines in Fig.~\ref{fig:noise_res_variance} is the largest and smallest form the same matrices.

\subsection{Uncorrelated noise}
Now let us consider the impact of uncorrelated additive and multiplicative noise inside reservoir. The results of numerical simulation for all three trained ESNs are given in Fig.~\ref{fig:noise_res_variance}(d--f). The points corresponding to additive noise are shown in purple, while the rest were prepared for multiplicative noise.

Equation (\ref{eq:var_noise_res}) transforms to
\begin{equation}\label{eq:var_uncorr_noise_res}
\begin{array}{c}
\mathrm{Var}[y^\mathrm{out}_t] \approx 
2D^U_A N\eta(\mathbf{W}^\mathrm{out}) + 2D^U_M\sum\limits_j\Big( W^\mathrm{out}_j \mathrm{E}[x^\mathrm{res}_{t,j}] \Big)^2 + \\
(1+2D^U_M)N^2\mu^2(\mathbf{W}^\mathrm{out})\cdot\overline{\mathrm{Var}[\mathbf{x}^\mathrm{res}_t]}
\end{array}
\end{equation}
for uncorrelated noise. As can be seen form Tab.~\ref{tab:W_out}, the values of $N\eta(\mathbf{W}^\mathrm{out})<N^2\mu^2(\mathbf{W}^\mathrm{out})$ for obtained after training $\mathbf{W}^\mathrm{out}$. The first multiplier is used for uncorrelated additive noise, while the multiplier with mean squared can be found before correlated additive noise. Therefore, the contribution of additive correlated noise is larger than uncorrelated additive noise if the noise intensities are the same. 

The impact of uncorrelated additive noise is rather easy to approximate. The minimal variance corresponds to only first term in (\ref{eq:var_uncorr_noise_res}), this level is shown by the black dashed lines in Fig.~\ref{fig:noise_res_variance}(d--f). 

Previously, the shapes of variances were very different for additive and multiplicative noise. Now for uncorrelated noise in reservoir they almost overlap (Fig.~\ref{fig:noise_res_variance}(d--f)). In order to explain this phenomenon, let us consider in more detail the second term in (\ref{eq:var_uncorr_noise_res}):
\begin{equation}\nonumber
\begin{array}{c}
2D^U_M\sum\limits_j\Big( W^\mathrm{out}_j \mathrm{E}[x^\mathrm{res}_{t,j}] \Big)^2 \approx 2D^U_M\eta(\mathbf{W}^\mathrm{out})\sum\limits_j \mathrm{E}^2[x^\mathrm{res}_{t,j}] \approx \\
2D^U_M\eta(\mathbf{W}^\mathrm{out})\sum\limits_j \big( f(u_t)  \big)^2.
\end{array}
\end{equation}
This is a very rough approximation, but if all input signals $u_t$ are positive as in our case, then the term $\big( f(u_t)  \big)^2\to 1$ for hyperbolic tangent used as $f()$. 
\begin{equation}\nonumber
2D^U_M\sum\limits_j\Big( W^\mathrm{out}_j \mathrm{E}[x^\mathrm{res}_{t,j}] \Big)^2 \approx 2D^U_M N\eta(\mathbf{W}^\mathrm{out}).
\end{equation}
If $D^U_A=D^U_M$, then it is equal to the first term for additive noise in (\ref{eq:var_uncorr_noise_res}).

As can be seen from Fig.~\ref{fig:noise_res_variance}, the uncorrelated noise is accumulated in reservoir a lot, and therefore the points of numerical simulation are located much higher than the black line. Moreover, the additional influence from the reservoir for correlated noise consisted of a constant increase in the variance of the output signal, but in the case of uncorrelated noise the contribution of the noisy reservoir is highly nonlinear and apparently depends on the type of input signal. 

\begin{figure}[h]
\includegraphics[width=\linewidth]{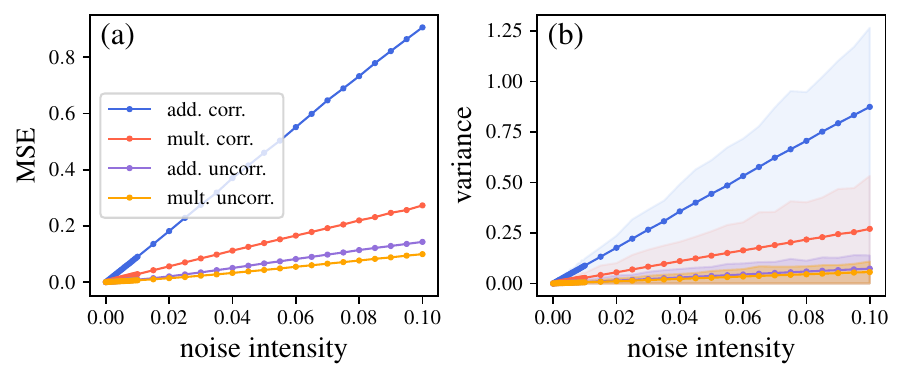}
\caption{Averaged MSE (a) and variance (b) of the output ESN signal $y^\mathrm{out}_t$ in the case of noise in the reservoir of trained ESNs with uniform $\mathbf{W}^\mathrm{res}$. Four noise sources were considered: correlated additive (blue) and multiplicative (red); uncorrelated additive (purple) and multiplicative (orange). Coloured backgrounds in (b) show the ranges of variance. }
\label{fig:noise_res_mse}
\end{figure}

Figure~\ref{fig:noise_res_mse} shows the dependences of MSE (a) and variance (b) on the noise intensity for one trained network with uniform matrix $\mathbf{W}^\mathrm{res}$ when correlated (blue and red points) and uncorrelated (purple and orange points) noise affect the reservoir. The line and points show the average level of variance and MSE depending on the noise intensity, while the background same color shows the spread of variance values. Variance and MSE almost overlap for additive and multiplicative uncorrelated noise. For correlated noise they diverge greatly. 

All dependences change almost linearly with increasing noise intensity. But for additive correlated noise, the variance and MSE grow faster. Therefore, additive correlated noise is the worst for this ESN. If one increase the noise intensity to $10^-2$, then even at this value the MSE for additive noise becomes close to $0.1$, which is already bad for the considered range of $u_t$ values. If the noise intensity increases further to $0.1$, then the linear growth of all dependencies remains.

As for multiplicative noise, correlated noise degrades network performance more than uncorrelated noise.

\section{Analytical prediction of the noise level}\label{sec:analytics}
In this paper we consider statistical methods for analysis of random signal and to predict the dispersion or variance of the output ESN signal $y^\mathrm{out}_t$. These methods have been already introduced and have been used successfully for deep neural networks. In this section we are using terms of variance of a random variable $\mathrm{Var}[Y]$ and its expected value $\mathrm{E}[Y]$. The variance can be calculated as $\mathrm{Var}[Y]=\mathrm{E}[Y^2]-\mathrm{E}^2[Y]$, where $\mathrm{E}^2[Y]=\big(\mathrm{E}[Y]\big)$. There are several properties of variance which will be used further:
\begin{itemize}
\item random variable $Y$ multiplied by a constant $c$: $\mathrm{Var}[cY]=c^2\mathrm{Var}[Y]$;
\item adding two random variables $Y_1$ and $Y_2$: $\mathrm{Var}[Y_1 + Y_2]=\mathrm{Var}[Y_1] + \mathrm{Var}[Y_2]$;
\item two multiplied uncorrelated random variables $Y_1$ and $Y_2$: $\mathrm{Var}[Y_1\cdot Y_2]=\mathrm{E}^2[Y_1]\mathrm{Var}[Y_2] + \Big(\mathrm{E}^2[Y_2]+\mathrm{Var}[Y_2]\Big)\mathrm{Var}[Y_1]$.
\end{itemize}

\subsection{Noise in output layer}\label{sec:analytics_output}
In the case of noise in the output layer, the noise operator is applied to the variable $x^\mathrm{out}_t$ in accordance with (\ref{eq:noise_out}). Then its variance can be calculated as
\begin{equation}\label{eq:an:out}
\begin{array}{c}
\mathrm{Var}[y^\mathrm{out}_t] =\mathrm{Var}\Big[ \sqrt{2D^C_A}\xi^{CA}_{t} \Big] + \mathrm{Var}\Big[ x^\mathrm{out}_{t}\cdot\big( 1+\sqrt{2D^C_M}\xi^{CM}_{t} \big) \Big] = \\
2D^C_A + 2D^C_M\cdot\mathrm{E}^2[x^\mathrm{out}_{t}] + (1+2D^C_M)\cdot\mathrm{Var}[x^\mathrm{out}_{t}]
\end{array}
\end{equation}
The variables $x^\mathrm{out}_t$ and $y^\mathrm{out}_t$ are the ESN output before and after the noise influence, respectively. All considered noise sources have zero mean, so the expected values for both variables are the same $\mathrm{E}[x^\mathrm{out}_{t}]=\mathrm{E}[y^\mathrm{out}_{t}]$. Therefore, (\ref{eq:an:out}) can be simply transformed to (\ref{eq:var_out}).

\subsection{Noise in reservoir}\label{sec:analytics_reservoir}
In the case when the noise is introduced into the reservoir, the variance of the output signal $y^\mathrm{out}_t=\sum\limits_{j=1}^N W^\mathrm{out}_j y^\mathrm{res}_{t,j}$ becomes
\begin{equation}\label{eq:an:res1}
\begin{array}{c}
\mathrm{Var}[y^\mathrm{out}_t] = \mathrm{Var}\Big[\sum\limits_{j=1}^N W^\mathrm{out}_j y^\mathrm{res}_{t,j}\Big] = \\

\mathrm{Var}\Big[\sqrt{2D^C_A}\xi^{CA}_t\cdot\sum\limits_{j=1}^N W^\mathrm{out}_j\Big] + \mathrm{Var}\Big[\sum\limits_{j=1}^N W^\mathrm{out}_j\sqrt{2D^U_A}\xi^{UA}_{t,j}\Big] + \\
\mathrm{Var}\Big[ \big( 1+ \sqrt{2D^C_M}\xi^{CM}_t\big) \sum\limits_{j=1}^N W^\mathrm{out}_j x^\mathrm{res}_{t,j}\big( 1+ \sqrt{2D^U_M}\xi^{UM}_{t,j} \big) \Big] = \\
\\
2D^C_A\Big( \sum\limits_{j=1}^N W^\mathrm{out}_j \Big)^2 + \sum\limits_{j=1}^N \Big(W^\mathrm{out}_j \Big)^2\cdot 2D^U_A + \\
2D^C_M \mathrm{E}^2\Big[ \sum\limits_{j=1}^N W^\mathrm{out}_j x^\mathrm{res}_{t,j}\big( 1+ \sqrt{2D^U_M}\xi^{UM}_{t,j} \big)\Big] + \\ (1+2D^C_M) \mathrm{Var}\Big[ \sum\limits_{j=1}^N W^\mathrm{out}_j x^\mathrm{res}_{t,j}\big( 1+ \sqrt{2D^U_M}\xi^{UM}_{t,j} \big)\Big] =\\
\\
2D^C_A\Big( \sum\limits_{j=1}^N W^\mathrm{out}_j \Big)^2 + 2D^U_A\sum\limits_{j=1}^N \Big(W^\mathrm{out}_j \Big)^2 + \\
2D^C_M \mathrm{E}^2\Big[ \sum\limits_{j=1}^N W^\mathrm{out}_j x^\mathrm{res}_{t,j}\Big] + \\
(1+2D^C_M) \sum\limits_{j=1}^N 2D^U_M\big(W^\mathrm{out}_j \mathrm{E}[x^\mathrm{res}_{t,j}]\big)^2 + \\
(1+2D^C_M) \sum\limits_{j=1}^N (1+2D^U_M)\Big(W^\mathrm{out}_j\Big)^2 \mathrm{Var}\Big[x^\mathrm{res}_{t,j}\Big].
\end{array}
\end{equation}
After introducing the designations of mean $\mu(\mathbf{W}^\mathrm{out})=\frac{1}{N}\sum\limits_{j=1}^N W^\mathrm{out}_j$ and mean of square $\eta(\mathbf{W}^\mathrm{out})=\frac{1}{N}\sum\limits_{j=1}^N \big(W^\mathrm{out}_j\big)^2$ for the output connection matrix, we get that
\begin{equation}\nonumber
\begin{array}{c}
\Big(\sum\limits_{j=1}^N W^\mathrm{out}_j \Big)^2 = N^2 \Big(\mu(\mathbf{W}^\mathrm{out})\Big)^2 = N^2 \mu^2(\mathbf{W}^\mathrm{out}) \\
\sum\limits_{j=1}^N \Big(W^\mathrm{out}_j \Big)^2 = N \eta(\mathbf{W}^\mathrm{out}) .
\end{array}
\end{equation}
Then (\ref{eq:an:res1}) transforms to
\begin{equation}\label{eq:an:res2}
\begin{array}{c}
\mathrm{Var}[y^\mathrm{out}_t] = 2D^C_A N^2\mu^2(\mathbf{W}^\mathrm{out}) + 2D^U_A N\eta(\mathbf{W}^\mathrm{out}) + \\
2D^C_M \mathrm{E}^2\Big[ y^\mathrm{out}_t] + 2D^U_M(1+2D^C_M) \sum\limits_{j=1}^N \big(W^\mathrm{out}_j \mathrm{E}[x^\mathrm{res}_{t,j}]\big)^2 + \\
(1+2D^C_M)(1+2D^U_M) \sum\limits_{j=1}^N \Big(W^\mathrm{out}_j\Big)^2 \mathrm{Var}\Big[x^\mathrm{res}_{t,j}\Big].
\end{array}
\end{equation}
After simple transform we get (\ref{eq:var_noise_res}).

\section{Self-closed ESN}\label{sec:closed_ESN}
In the previous sections, we have considered not self-closed ESNs. This means, that the input signal $u_t$ was set from the original Mackey-Glass implementation. Now, we are going to look what happens with the output signal if ESN is self-closed. This means that ESN's output at time $t$ is used as the input for ESN at the next time $u_{t+1}$. In this case, there is a noise impact not only on individual parts of the network such as a reservoir or output layer, but also a noisy signal is supplied indirectly to the network input. Naturally, this will lead to the fact that at some point the network will completely lose the ability to predict and will only produce a noisy signal. Let us consider what types of noise in this case are the most critical and with what noise intensities the network continues to work.

\begin{figure}[h]
\includegraphics[width=\linewidth]{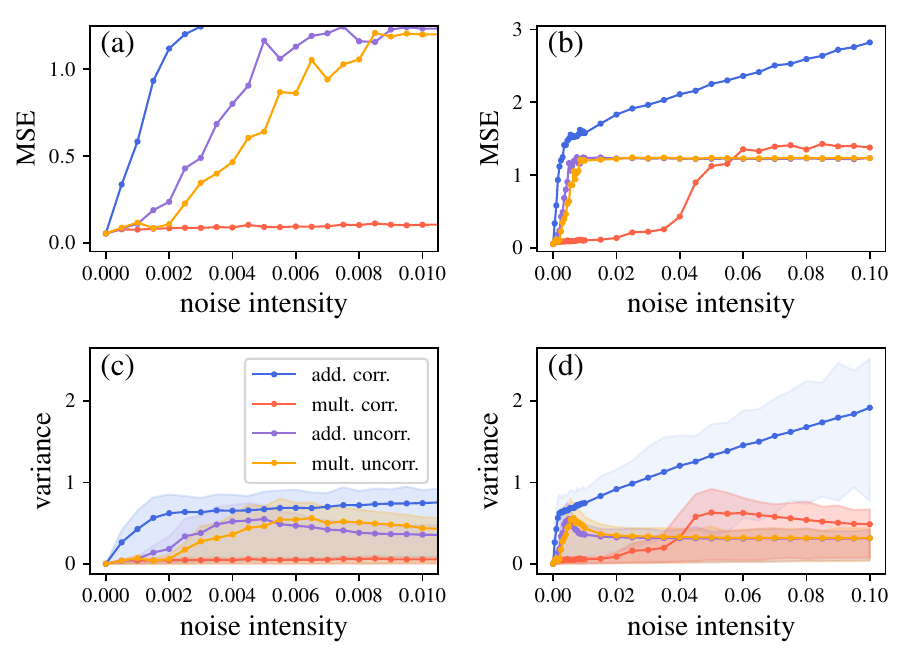}
\caption{Averaged MSE (a) and variance (b) of the output of self-closed ESN in the case of noise in the reservoir of trained ESNs with uniform $\mathbf{W}^\mathrm{res}$. Four noise sources were considered: correlated additive (blue) and multiplicative (red); uncorrelated additive (purple) and multiplicative (orange). Coloured backgrounds in (b) show the ranges of variance.}
\label{fig:noise_free_res_mse}
\end{figure}

Figure~\ref{fig:noise_free_res_mse} shows how the MSE and variance of the ESN output signal changes if noise is applied to the network reservoir. The error is shown in the upper panels for two scales of noise intensity $D\in[0; 0.01]$ (a) and $D\in[0; 0.1]$ (b). As can be seen from the graphs for a large scale, at some high noise intensities, almost all dependences stop growing, and MSE remains almost unchanged. This saturation value can be taken to mean that noise is no longer as critical to the network. However, if one pays attention to the MSE values to which this corresponds, it becomes clear that in this case the useful signal is completely lost, and only noise is observed at the network output.

\begin{figure}[h]
\includegraphics[width=\linewidth]{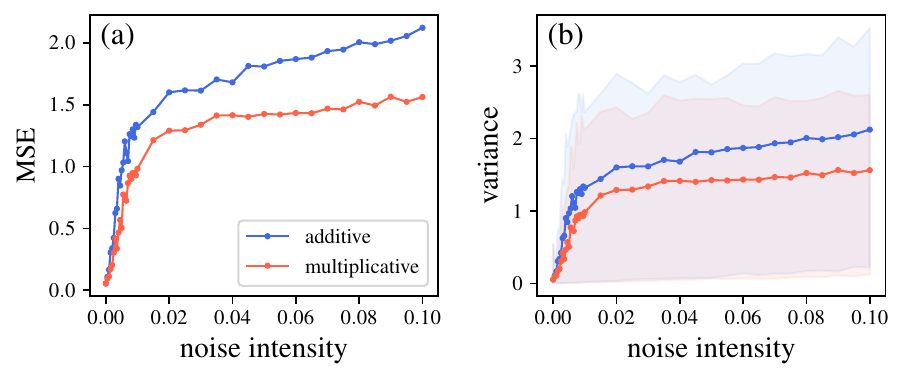}
\caption{Averaged MSE (a) and variance (b) of the output of self-closed ESN in the case of noise in the output layer of trained ESNs with uniform $\mathbf{W}^\mathrm{res}$. Two noise sources were considered: additive (blue) and multiplicative (red). Coloured backgrounds in (b) show the ranges of variance.}
\label{fig:noise_free_out_mse}
\end{figure}

The only exception is additive correlated noise, for which MSE continues to grow linearly. This type of noise is generally the worst noise (Fig.~\ref{fig:noise_free_res_mse}(a,b), blue). Almost immediately, at very low noise intensities, MSE approaches 1. Such a value of MSE for the initial range of values $u_t\in[0.4;1.4]$ indicates a complete loss of the signal and that we then observe simply noise, which variance is controlled by the intensity $D^C_A$, so the error keeps growing.

As for other types of noise, correlated multiplicative noise performed best (Fig.~\ref{fig:noise_free_res_mse}(a,b), red). Its critical impact is observed at $D^C_M>0.05$, which makes it the last type of noise to affect performance. Uncorrelated noise sources are very similar to each other (see Fig.~\ref{fig:noise_free_res_mse}(a,b), blue and orange), but the negative effect of additive noise appears at lower noise intensities. The critical intensities are the following: $D^U_A=2.5\cdot 10^{-3}$, $D^U_M=4\cdot 10^{-3}$. Here, by critically important intensities we mean noise intensities for which MSE exceeds $0.5$, which corresponds to a complete loss of useful information, since the range of initial $u_t$ was from $0.4$ to $1.3$. The lower fragments of Fig.~\ref{fig:noise_free_res_mse}(c,d) show the corresponding distributions of variances for the corresponding types of noise from the upper panels. These panels fully confirm the results for MSE.

Figure~\ref{fig:noise_free_out_mse} shows similar dependencies of MSE and variance but for noise in the output layer. For not self-closed ESN, the impact of noise in the output layer was the most trouble-free compared to a reservoir. In self-closed ESN, additive and multiplicative noise in the output layers become noise in the input signal. Therefore, even small noise intensities lead to a significant increase in MSE and variance. Both dependencies continue to grow for larger noise intensities but make it slower.

\section{Other activation functions}\label{sec:activation_funciton}
The main goal of this work is to provide a general evaluation of the impact of noise on trained ESNs. Hyperbolic tangent is often used for reservoir computing, but it is important to understand what can be achieved with other activation functions for generality. In our previous works, we have already separately considered how different activation functions can affect the propagation of noise in simplified ESNs. In papers \cite{Semenova2023ND} and \cite{Semenova2023PND}, we have considered a linear activation function, sigmoid and hyperbolic tangent, but for an untrained and very simplified network. It is important to underline, that trained networks uncovered many aspects that were not possible to take into account for simplified networks. Therefore, we applied all the studies that were described in the previous sections to other trained networks that had reservoir neurons with a linear $f(x)=x$ and sigmoid $f(x)=1/(1+e^{-x})$ activation functions.

To begin with, it should be noted that if noise is introduced into the output layer, the effect of the noise will be the same as described in Sec.~\ref{sec:noise_output_layer} regardless of the reservoir activation function. Therefore, further we will focus more on the specific influence of noise in the reservoir.

As for the reservoir, the primary influence on the result is not the activation function itself, but the connection matrices with which the network can be considered trained. We considered all three types of reservoir matrix $\mathbf{W}^\mathrm{res}$: uniform, band 5\% and band 10\%. In the tables \ref{tab:matrices_sigmoid} and \ref{tab:matrices_linear} we show the statistics of all matrices for the sigmoid and linear activation function. The results were similar for all three types, so further in the figures we will show only data for random uniform $\mathbf{W}^\mathrm{res}$.

\begin{table}[t]\begin{center}
\begin{tabular}{| c | c | c | c |}
\hline
$\mathbf{W}^\mathrm{res}$ & uniform & band 5\% & band 10\%  \\ \hline
$\mu(\mathbf{W}^\mathrm{res})$   & $-6.76\cdot 10^{-5}$ & $-0.0010$           & $-0.0004$ \\
$\mu^2(\mathbf{W}^\mathrm{res})$ & $4.57\cdot 10^{-9}$  & $9.58\cdot 10^{-7}$ & $1.8071\cdot 10^{-7}$ \\
$\eta(\mathbf{W}^\mathrm{res})$  & $0.2331$             & $0.2980$            & $0.3355$ \\
std$(\mathbf{W}^\mathrm{res})$   & $0.48284229$         & $0.5459$            & $0.5792$ \\
spectral radius, $s$             & 11                   & 14                  & 13.6 \\
\hline
\hline
$\mu(\mathbf{W}^\mathrm{out})$   &  $0.0031$        & $0.0031$        & $0.0031$ \\
$\mu^2(\mathbf{W}^\mathrm{out})$ & $9.39\cdot 10^{-6}$  & $9.39\cdot 10^{-6}$ & $9.39\cdot 10^{-6}$ \\
$\eta(\mathbf{W}^\mathrm{out})$  & $2.77\cdot 10^{10}$  & $1.11\cdot 10^{11}$ & $3.36\cdot 10^{11}$ \\
std$(\mathbf{W}^\mathrm{out})$   & $1.66\cdot 10^5$     &  $3.33\cdot 10^5$   & $5.79\cdot 10^5$ \\
\hline
\hline
testing MSE                      & $3.91\cdot 10^{-5}$  & $2.64\cdot 10^{-4}$ & $0.0011$ \\
self-closed MSE                  & $0.0387$             & $0.0577$            & $0.0737$ \\
\hline
\end{tabular}
\end{center}
\caption{Statistical characteristics of matrices $\mathbf{W}^\mathrm{res}$ and $\mathbf{W}^\mathrm{out}$ of three trained ESNs with sigmoid activation function.}
\label{tab:matrices_sigmoid}
\end{table}

\begin{table}[t]\begin{center}
\begin{tabular}{| c | c | c | c |}
\hline
$\mathbf{W}^\mathrm{res}$ & uniform & band 5\% & band 10\%  \\ \hline
$\mu(\mathbf{W}^\mathrm{res})$   & $-2.89\cdot 10^{-5}$ &   $-0.0002$        & $-3.33\cdot 10^{-5}$ \\
$\mu^2(\mathbf{W}^\mathrm{res})$ & $8.36\cdot 10^{-10}$ & $3.71\cdot 10{-8}$ &  $1.11\cdot 10^{-9}$ \\
$\eta(\mathbf{W}^\mathrm{res})$  & $0.0019$             & $0.0016$           & $0.0019$ \\
std$(\mathbf{W}^\mathrm{res})$   & $0.0437$             & $0.0405$           & $0.0434$ \\
spectral radius, $s$             & 1                    & 1                  & 1 \\
\hline
\hline
$\mu(\mathbf{W}^\mathrm{out})$   & $0.0030$            & $0.0033$            & $0.0029$ \\
$\mu^2(\mathbf{W}^\mathrm{out})$ & $8.88\cdot 10^{-6}$ & $1.06\cdot 10^{-5}$ & $8.38\cdot 10^{-6}$ \\
$\eta(\mathbf{W}^\mathrm{out})$  & $6.01\cdot 10^{17}$ & $2.50\cdot 10^{12}$ & $8.74\cdot 10^{17}$ \\
std$(\mathbf{W}^\mathrm{out})$   & $7.75\cdot 10^8$    & $1.58\cdot 10^6$    & $9.35\cdot 10^8$ \\
\hline
\hline
testing MSE                  & $1.36\cdot 10{-5}$     & $1.71\cdot 10{-5}$   & $1.44\cdot 10^{-5}$ \\
self-closed MSE              & $0.0137$               & $0.0059$             & $0.0061$ \\
\hline
\end{tabular}
\end{center}
\caption{Statistical characteristics of matrices $\mathbf{W}^\mathrm{res}$ and $\mathbf{W}^\mathrm{out}$ of three trained ESNs with linear activation function.}
\label{tab:matrices_linear}
\end{table}

\begin{figure*}[h]
\includegraphics[width=\linewidth]{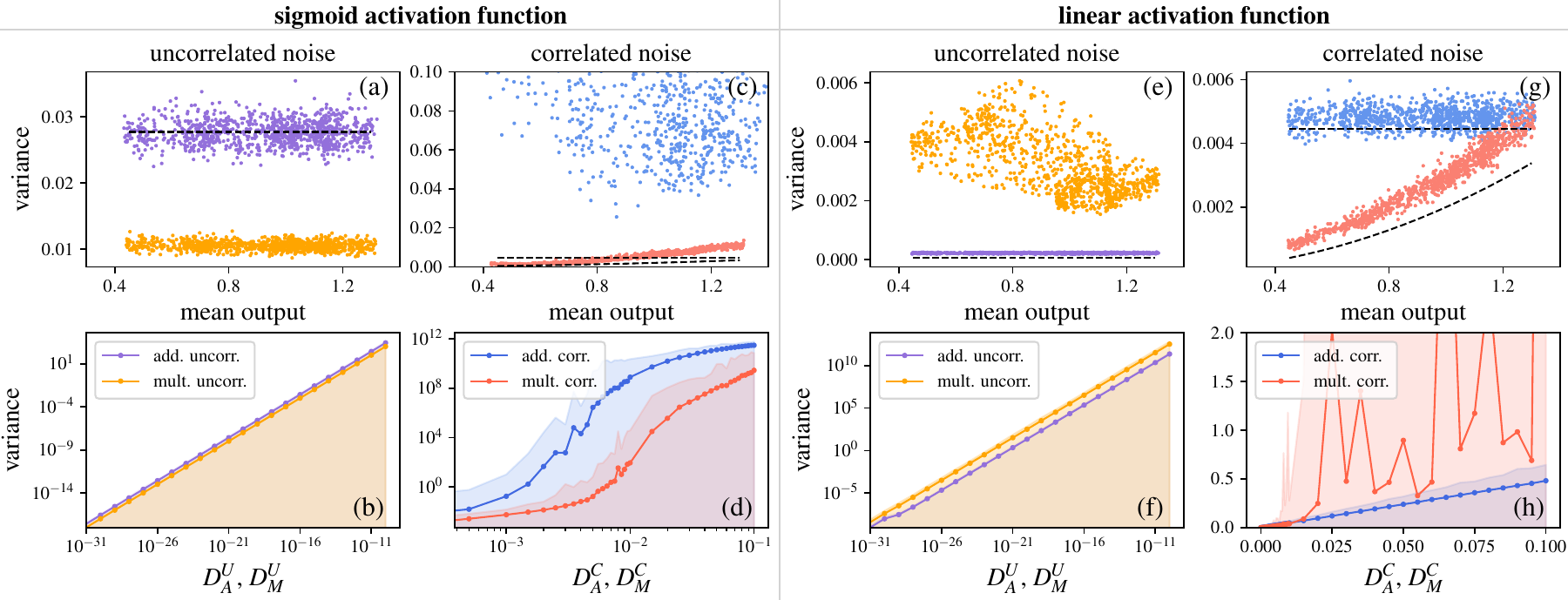}
\caption{The impact of activation functions on noise accumulation in reservoir. Parameters: $D^U_A=D^U_M=10^{-25}$ for linear activation function, $D^U_A=D^U_M=10^{-15}$ for sigmoid activation function, $D^C_A=D^C_M=10^{-3}$ for both cases.}
\label{fig:activation function}
\end{figure*}

As shown in tables, all the considered trained networks have a very strange feature. The values of $\eta(\mathbf{W}^\mathrm{out})$ are very large. As shown in previous sections, this value is a multiplier for the variance corresponding to uncorrelated noise. Therefore, for such networks with such a feature of the output connection matrix, uncorrelated noise is very critical. To see how the variances depend on the mean output signal (Fig.~\ref{fig:activation function}), we have to consider noise intensities that are much lower than before. For the previously considered noise intensity of $10^{-3}$, all these dependencies go into such large ranges variance and error that these networks are no longer able to predict anything.

Fig.~\ref{fig:activation function} shows the dependence of the variation on the average output signal for uncorrelated noise (a,e) and correlated noise (c,g). Fig.~\ref{fig:activation function} is divided into two parts by a vertical gray line. All graphs on the left were obtained for the sigmoid, and the graphs on the right were obtained for the linear activation function. For correlated noise, the usual intensities $D^C_A=D^C_M=10^{-3}$ were used, and for uncorrelated noise, much lower values were considered, namely $D^U_A=D^U_M=10^{-15}$ for the sigmoid and $D^U_A=D^U_M=10^{-25}$ for the linear function. For correlated additive and both uncorrelated noise types, the black dashed lines show analytical prediction of the noise level based on (\ref{eq:var_corr_noise_res}) and (\ref{eq:var_uncorr_noise_res}), and they are in good agreement with the results of numerical simulations. Where modelling and analytics diverge greatly probably indicates that noise is heavily accumulating in the reservoir. This can be seen for the correlated multiplicative noise in Fig.~\ref{fig:activation function}(e) for sigmoid function.

The lower panels of Fig.~\ref{fig:activation function} show the dependencies of the average variance on the noise intensity. For uncorrelated noise, we had to reduce the considered range to $[10^{-30},10^{-10}]$. But even for such small intensities the variance becomes more than $10$ for sigmoid and more than $10^{10}$ for a linear activation function. For the sigmoid, as was the case for the hyperbolic tangent, additive noise (both correlated and uncorrelated) is more critical for the network than multiplicative noise. The dependences for additive noise are located above in Fig.~\ref{fig:activation function}(b,d). However, if one look at similar graphs for a linear function, it is quite the opposite, the multiplicative noise is more critical. This is most likely due to the fact that the sigmoid and hyperbolic tangent are bounded functions with maximum limit equal to unity, while the linear function has no such restrictions.

\section*{Conclusion}
\textbf{Output connection matrices.} From the point of view of noise accumulation, the most significant are the statistical characteristics of the connection matrix $\mathbf{W}^\mathrm{out}$, which is obtained as a result of ESN training. The main characteristics are the average of the output matrix multiplied by the number of neurons in the reservoir and then squared $N^2\mu^2(\mathbf{W}^\mathrm{out})$ and the average square of the same matrix multiplied by the number of neurons in reservoir $N\eta(\mathbf{W}^\mathrm{out})$. The presence of correlated noise in the variance of the output signal depends on the value $N^2\mu^2(\mathbf{W}^\mathrm{out})$, while the value $N\eta(\mathbf{W}^\mathrm{out})$ is responsible for uncorrelated noise. If trained ESN has $N\eta(\mathbf{W}^\mathrm{out})>1$, then in such a network uncorrelated noise accumulates very quickly, and this leads to the fact that the variance becomes incomparably large even for a weak noise. Sometimes the intensity of $10^{-20}$ is enough to completely lose the useful signal. Similarly, $N^2\mu^2(\mathbf{W}^\mathrm{out})>1$ leads to uncontrolled accumulation of additive correlated noise.

It is not always possible to train ESN in such a way that the required multipliers $N^2\mu^2(\mathbf{W}^\mathrm{out})$ and $N\eta(\mathbf{W}^\mathrm{out} )$ met the above conditions $N\eta(\mathbf{W}^\mathrm{out})<1$, and uncorrelated noise accumulates very strongly in such networks. Judging by preliminary analysis, this problem can be solved using the bias offset.

\textbf{Reservoir connection matrices.} This paper considered three commonly used types of reservoir connection matrices $\mathbf{W}^\mathrm{res}$: random uniform and two band matrices with 5\% connectivity and 10\% connectivity. In terms of noise accumulation, the results for these three types are not much different.

\textbf{Analytics.} This paper proposes a rough estimate of the ESN output variance for all four noise types. In this analysis, we are currently ignoring the issues of noise accumulation in the reservoir. A more accurate estimate must first take into account the characteristics of the input signal, and most likely can only be achieved using numerical simulation. Perhaps this will be considered in future works.

\textbf{Noise in the output layer.} In this paper, all variance and MSE estimation methods have been described in detail for three trained ESNs with hyperbolic tangent as the activation function in the reservoir. We have shown that additive and multiplicative noise in the output layer do not greatly affect the variance and error of the output signal. Even at high intensities $D=0.1$ this leads to a quite adequate error $MSE\approx 0.2$. The variance of the output signal in this case agrees very well with the analytical estimates (\ref{eq:var_out}). The situation changes greatly when the network becomes self-closed. In this case, the noise at the output of the ESN is simultaneously the noise at the input of the same ESN at the next time moment, and both types of noise begin to accumulate strongly. Additive noise accumulates faster than multiplicative noise and already $D=0.005$ can be critical.

\textbf{Noise in the reservoir.} As for the noise inside the reservoir, here we looked at all four types of noise. Depending on how the noise affects each neuron, the noise was additive or multiplicative, and depending on how it affects the entire reservoir, it was correlated or uncorrelated noise. Correlated noise results in variances visually similar to those of the output layer. For uncorrelated noise, the position of the points and the nonlinearity of the noise accumulation were more pronounced. The variances for additive and multiplicative noise practically overlapped each other. As for the variance values themselves, at the same intensity $10^{-3}$ the variance and error for correlated noise was greater than for uncorrelated noise. This is explained by the fact that for all three trained networks $N^2\mu^2(\mathbf{W}^\mathrm{out})>N\eta(\mathbf{W}^\mathrm{res})$. The dependence of variance and error on noise intensity is linearly increasing for all four types of noise, but for correlated noise the increase was faster, especially for correlated additive noise. If the system is self-closed, then this leads to an unexpected result. Additive correlated noise remains the worst noise for ESNs, while the best noise is correlated multiplicative noise.

\textbf{Activation function.} Next, we considered several ESNs with other activation functions: sigmoid and linear. It has been shown that the activation function has no effect on the variance if noise is introduced into the output layer. However, if the noise is fed into the reservoir, the difference is very significant. Since the goal of this work was to examine how noise affects the trained ESN, it is not possible to control the statistical properties of the output connection matrix $\mathbf{W}^\mathrm{out}$. For all three types of echo networks with a sigmoid activation function, it was found that after training the value of $N\eta(\mathbf{W}^\mathrm{out})\gg 1$, therefore uncorrelated noise accumulates in them inadequately quickly. This is fully consistent with our assumptions when considering ESNs with hyperbolic tangent, but we have not previously dealt with such large $\eta(\mathbf{W}^\mathrm{out})$. As for correlated noise, the results are similar to what was obtained for the hyperbolic tangent. For sigmoid function, additive (both correlated and uncorrelated) noise accumulates slightly faster than multiplicative noise. The same result was obtained for the hyperbolic tangent.

For a linear activation function, all trained ESNs had an even larger $\eta(\mathbf{W}^\mathrm{out})$, and their uncorrelated noise accumulated even faster. Therefore, noise with an intensity of $10^{-10}$ led to a variance greater than $10^{10}$ and a comparable error. For such networks, even small uncorrelated noise is very critical. For correlated noise the situation is better, but if previously correlated additive noise was more critical than multiplicative noise, here the situation is the opposite. This is most likely due to the fact that sigmoid and hyperbolic tangent are bounded functions with maximum limit 1, but for a linear function there are no such restrictions.

\section*{Data availability}
The data that support the findings of this study are available from the corresponding author upon reasonable request.

\section*{Acknowledgements}
This work was supported by the Russian Science Foundation (project No. 23-72-01094) https://rscf.ru/project/23-72-01094/ 

%
%

\end{document}